\documentclass{article}

\usepackage{PRIMEarxiv}
\usepackage[utf8]{inputenc}
\usepackage[T1]{fontenc}
\usepackage{hyperref}
\usepackage{url}
\usepackage{booktabs}
\usepackage{amsfonts}
\usepackage{nicefrac}
\usepackage{microtype}
\usepackage{amsmath}
\usepackage{graphicx}
\graphicspath{{media/}}
\usepackage{subcaption}
\usepackage[inline]{enumitem}
\usepackage{algorithm}
\usepackage{algpseudocode}
\usepackage{rotating}
\usepackage{multirow}
\usepackage{makecell}
\usepackage{threeparttable}

\title{HiMX: Mixed-Format Matrix Multiplication Based on Customized HiFloat4 Representations}

\author{
Zhiyuan Zhang, Yanzhao Li, Hui Dong, Sifan Cheng, Jingwei Zhang \\
Huawei Technology \\
\texttt{\{zhangzhiyuan45, liyanzhao\}@huawei.com}
}

\begin{document}

\maketitle


\begin{abstract}

4-bit weight-activation quantization (W4A4) is an important approach for efficient large language model (LLM) inference. Recent hierarchical formats such as HiFloat4 (HIF4) improve the representational range of 4-bit quantization through multi-level scaling. However, the same local mantissa representation in HIF4 may not be equally suitable for weights and activations, which exhibit distinct distribution characteristics.

To address this limitation, we propose \textbf{HiMX}, a mixed-mantissa framework built upon the hierarchical scaling structure of HIF4. Our key observation is that the two HIF4-compatible mantissa representations, S1P2 and E2M1, exhibit complementary quantization behavior: the denser S1P2 format provides higher resolution for concentrated weight distributions, while the wider-spaced E2M1 format provides complementary quantization characteristics for activation tensors. HiMX enables operand-tailored mantissa representations while converting heterogeneous mantissas into fixed-point integer operands with operand-specific widths, allowing different operand formats to share a unified integer multiply-accumulate (MAC) array.

Block-level analysis on Qwen3-8B shows that HIF4-S1P2 is better suited for weights while HIF4-E2M1 is better suited for activations, and their mixed-format combination achieves lower output MSE than homogeneous configurations. Across multiple benchmarks, HiMX reduces the average end-to-end accuracy loss from 2.38 to 1.77 points on Qwen3-8B and from 2.23 to 1.85 points on Qwen3-30B-A3B, demonstrating the potential benefit of heterogeneous mantissa representations for W4A4 LLM inference.

\end{abstract}

\section{Introduction}
\label{sec:intro}

4-bit quantization has become an important direction for efficient large language model (LLM) inference\cite{gptq,awq,smoothquant,zeroquant}, with recent formats such as MXFP4, NVFP4, and HiFloat4 (HIF4) exploring different trade-offs between numerical accuracy\cite{mxfp4,nvfp4,hif4,ocp_mx,int_vs_fp}, representation efficiency, and hardware cost. Among them, HIF4 introduces a three-level scaling hierarchy to improve the utilization of the limited 4-bit representation space while capturing both inter-group and intra-group dynamic ranges\cite{hif4,hifa4}. Its large 64-element group size further enables matrix multiplication to be implemented in a highly fixed-point manner, reducing hardware area and power consumption\cite{hif4}. These properties make HIF4 an attractive foundation for efficient W4A4 inference\cite{hif4, OSC}.

To investigate whether further gains can be obtained from operand-aware representations, we first characterize the local numerical distributions of LLM weights and activations. Specifically, we analyze the local 4-element dynamic range and find that activation tensors exhibit a modest but consistent shift toward higher local dynamic-range distribution than weights. This observation suggests that the two operands may have different requirements for their local 4-bit representations: weights may benefit more from a denser format, while activations may benefit from a representation with wider numerical coverage.

Motivated by this observation, we further investigate the local mantissa representations available within the HIF4 framework. HIF4-compatible S1P2 and E2M1 provide different trade-offs between local resolution and dynamic range\cite{hif4,ocp_mx}. S1P2 offers denser representation levels, whereas E2M1 provides a wider dynamic range\cite{ocp_mx} and is also widely adopted in recent 4-bit floating-point formats such as MXFP4 and NVFP4 \cite{mxfp4,nvfp4}. Our analysis shows that these two mantissa representations exhibit complementary quantization behavior across weights and activations, motivating an operand-dependent selection of local mantissa formats while retaining the hierarchical scaling structure of HIF4.

Recent work has also explored heterogeneous numerical representations\cite{mant,adamx} for different operands. In particular, AdaMX investigates representation heterogeneity across blocks and operands, demonstrating the potential of adapting low-precision formats to tensor characteristics \cite{adamx}. HiMX is complementary to this line of work but focuses specifically on heterogeneous mantissa representations within the HIF4 hierarchical scaling framework. Instead of introducing independent numerical formats for different operands, HiMX retains the HIF4 scaling hierarchy and 4-bit structure while selecting between HIF4-compatible mantissa formats according to tensor-dependent characteristics\cite{mant,adamx,m2xfp}. Moreover, HiMX adopts a unified integer computation path. Heterogeneous mantissas are converted into fixed-point integer operands with operand-specific widths and aligned at the multiplier input, allowing different operand formats to share the same integer MAC array. This design is inspired by prior microscaling and shared-microexponent architectures that exploit integer-oriented or shared-scale computation \cite{mxfp4,msfp,shared_microexp,m2xfp}.

Based on these observations, we propose \textbf{HiMX}, a mixed-mantissa framework for HIF4 quantization. In our main configuration, HiMX employs HIF4-S1P2 for weights and HIF4-E2M1 for activations. By combining operand-dependent mantissa selection with a unified integer computation datapath, HiMX improves quantization fidelity while preserving the hardware-friendly characteristics of HIF4.

Our experiments on Qwen3 models demonstrate that heterogeneous mantissa representations can improve both quantization fidelity and end-to-end accuracy\cite{qwen3_report}. On Qwen3-8B, HIF4-S1P2 achieves lower weight quantization error, while activation tensors show a preference for HIF4-E2M1; their mixed-format combination further achieves lower GEMM output MSE than the evaluated homogeneous configurations. Across multiple benchmarks, HiMX reduces the average end-to-end accuracy loss from 2.38 to 1.77 percentage points on Qwen3-8B and from 2.23 to 1.85 percentage points on Qwen3-30B-A3B.

\noindent\textbf{Contributions.}
The main contributions of this work are summarized as follows:

\begin{itemize}

\item \textbf{Local Distribution Characterization:}
We analyze the local 4-element dynamic-range distributions of LLM weights and activations and observe different dynamic-range characteristics between the two operands.

\item \textbf{Mixed-Mantissa HIF4 Quantization:}
We propose \textbf{HiMX}, which extends the HIF4 hierarchical scaling framework with operand-dependent local mantissa representations, using HIF4-S1P2 for weights and HIF4-E2M1 for activations.

\item \textbf{Unified Mixed-Format Computation:}
We introduce a \textbf{format-to-fixed-point mapping} and width-alignment mechanism that converts heterogeneous mantissa formats into fixed-point integer operands with operand-specific widths, enabling different HIF4 representations to share a unified integer MAC array.

\item \textbf{End-to-End Evaluation:}
We evaluate HiMX on Qwen3 models across multiple benchmarks and demonstrate improved accuracy retention over unified HIF4 quantization with a small additional hardware cost.

\end{itemize}

\section{Characterization of Weight-Activation Distribution Heterogeneity}
\label{sec:analysis}

To establish the rationale for heterogeneous mantissa design within a unified 3-level microscaling quantization framework\cite{hif4}, this section analyzes the quantization behavior of HIF4-S1P2 and HIF4-E2M1 on LLM weights and activations. First, the respective quantization procedures are formalized under the shared HIF4 hierarchy. Next, we profile the post-scaling local dynamic ranges\cite{microscopiq} of four-element groups to characterize distribution differences between operands. Finally, we evaluate quantization errors at both tensor and GEMM levels to examine whether these distribution differences translate into different mantissa preferences.

\subsection{HIF4-S1P2 and HIF4-E2M1 Quantization Architecture}
\label{sec:hif4_quantization}

\begin{figure}[t]
\centering
\begin{subfigure}[b]{0.48\linewidth}
    \centering
    \includegraphics[width=\linewidth]{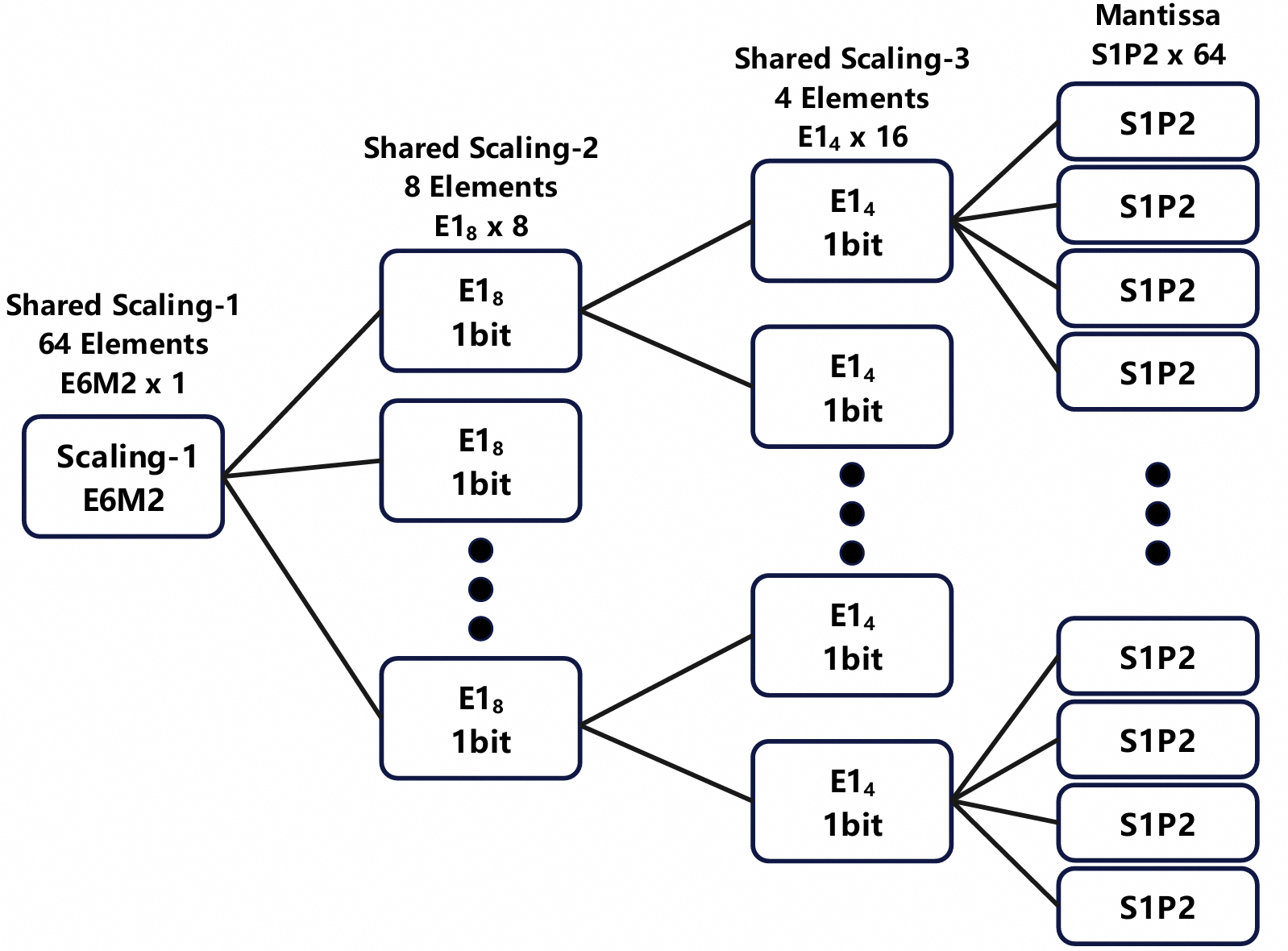}
    \caption{HIF4-S1P2 Scale Hierarchy\cite{hif4}.}
    \label{fig:scale_hif4_s1p2}
\end{subfigure}
\hfill
\begin{subfigure}[b]{0.48\linewidth}
    \centering
    \includegraphics[width=\linewidth]{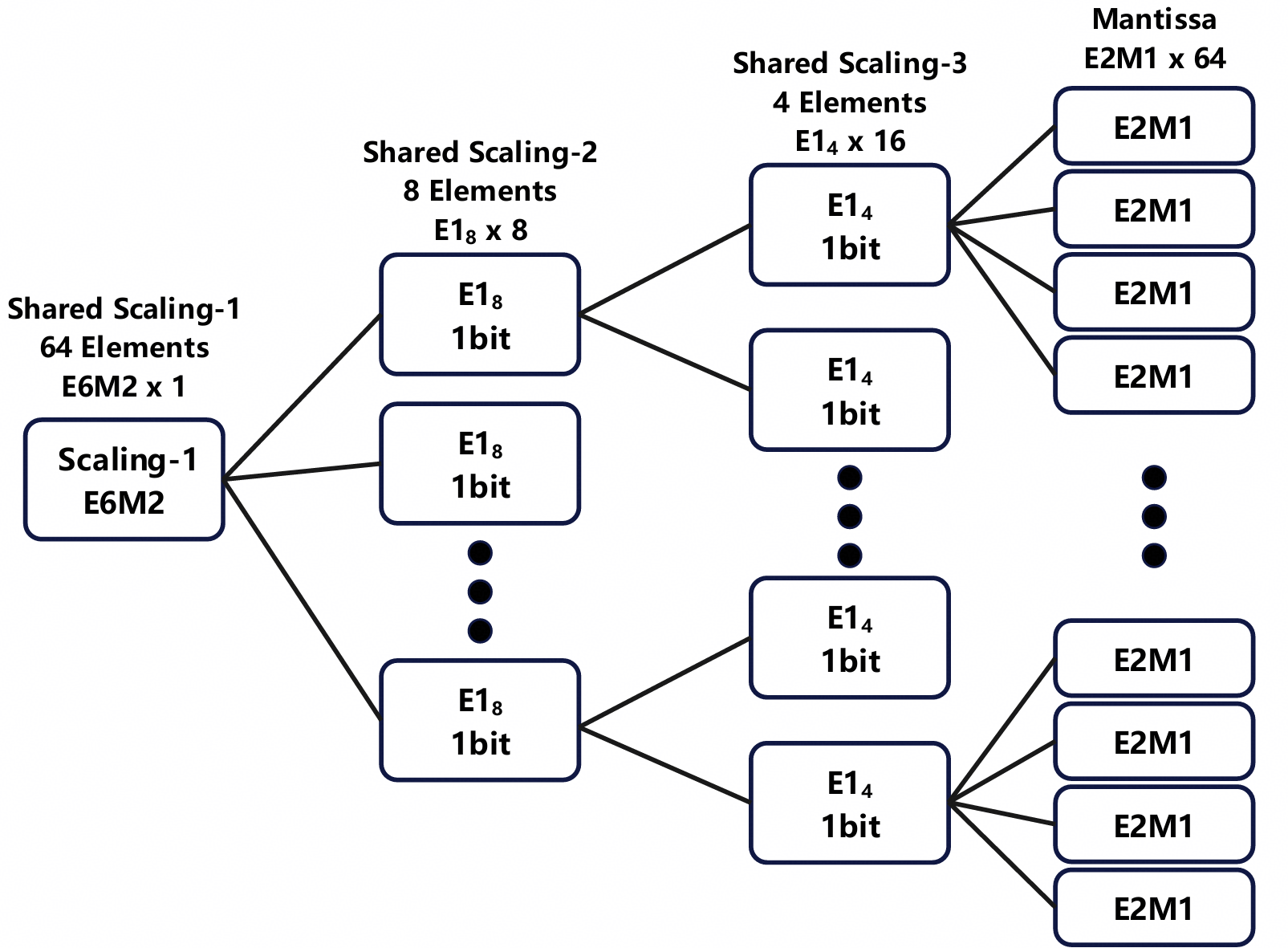}
    \caption{HIF4-E2M1 Scale Hierarchy.}
    \label{fig:scale_hif4_e2m1}
\end{subfigure}
\caption{Overview of the proposed HIF4 3-level microscaling\cite{hif4} quantization architecture for (a) S1P2 and (b) E2M1 variants. A 64-element block shares an E6M2 global scale factor ($\mathrm{SF}$), augmented by eight 1-bit mid-level scale extensions ($\mathrm{E1}_8$) and sixteen 1-bit fine-level scale extensions ($\mathrm{E1}_4$) to capture local magnitude variations prior to element quantization.}
\label{fig:scale_hif4}
\end{figure}

\begin{algorithm}[t]
\caption{HIF4 Block Quantization and Dequantization}
\label{alg:hif4_quant}
\begin{algorithmic}[1]
\Require 64-element block $V_{64} = \{v_1, \ldots, v_{64}\}$, format selection $\mathcal{F} \in \{\text{S1P2}, \text{E2M1}\}$
\Ensure Quantized-dequantized block $\hat{V}_{64}$
\State Set hyper-parameters $(\alpha, \beta_1, \beta_2)$ according to format $\mathcal{F}$
\For{$i = 1$ \textbf{to} $16$} \Comment{Extract 16 4-element local peaks}
    \State $V_4[i] \gets \max_{j \in \{4i-3, \ldots, 4i\}} |v_j|$
\EndFor
\For{$k = 1$ \textbf{to} $8$} \Comment{Extract 8 8-element mid-level peaks}
    \State $V_8[k] \gets \max(V_4[2k-1], V_4[2k])$
\EndFor
\State $V_{\max} \gets \max_{k} V_8[k]$ \Comment{Global block peak}
\State $\mathrm{SF} \gets \mathrm{clamp}(V_{\max} / \alpha, 2^{-48}, 49152)$
\State $\mathrm{E6M2} \gets \mathrm{Quantize}_{\mathrm{E6M2}}(\mathrm{SF})$ \Comment{Level-1 global scale}
\For{$k = 1$ \textbf{to} $8$} \Comment{Level-2 1-bit scales}
    \State $\mathrm{E1}_8[k] \gets \mathbb{I}(V_8[k] / \mathrm{E6M2} \ge \beta_1)$
\EndFor
\For{$i = 1$ \textbf{to} $16$} \Comment{Level-3 1-bit scales}
    \State $\mathrm{E1}_4[i] \gets \mathbb{I}(V_4[i] / (\mathrm{E6M2} \cdot 2^{\mathrm{E1}_8[\lceil i/2 \rceil]}) \ge \beta_2)$
\EndFor
\For{$j = 1$ \textbf{to} $64$} \Comment{Element-wise scaling \& element quantization}
    \State $i \gets \lceil j / 4 \rceil$, \quad $k \gets \lceil j / 8 \rceil$
    \State $S_{\text{HIF4}}(j) \gets \mathrm{E6M2} \cdot 2^{\mathrm{E1}_8[k] + \mathrm{E1}_4[i]}$
    \State $z_j \gets |v_j| / S_{\text{HIF4}}(j)$
    \State $\hat{v}_j \gets \mathrm{sign}(v_j) \cdot S_{\text{HIF4}}(j) \cdot \mathrm{Quantize}_{\mathcal{F}}(z_j)$
\EndFor
\State \Return $\hat{V}_{64} = \{\hat{v}_1, \ldots, \hat{v}_{64}\}$
\end{algorithmic}
\end{algorithm}

To systematically evaluate the impact of local mantissa representations under our three-level scale hierarchy, we introduce two representative 4-bit variants: \textbf{HIF4-S1P2} and \textbf{HIF4-E2M1}. The \textbf{E2M1} format corresponds to the FP4 element format specified by OCP MX, while S1P2 is the fixed-point local format used by HIF4. We therefore use the two formats under the same HIF4 hierarchical scaling structure to isolate the effect of local format selection from the effect of hierarchical scaling. Specifically, the \textbf{S1P2} format adopts the sign-magnitude fixed-point representation used as the local format in HIF4, where S denotes the sign bit, the 1 before P indicates a 1-bit integer part, and the 2 after P indicates a 2-bit fractional part (P is the binary point).The \textbf{E2M1} format corresponds to the 4-bit floating-point element format defined in the OCP MXFP4 specification, with one sign bit, two exponent bits, and one mantissa bit\cite{ocp_mx}.

In fine-grained micro-scaling formats, in-group quantization faces an inherent trade-off between local precision and dynamic range\cite{mxfp4,nvfp4,nxfp,int_vs_fp}. S1P2 offers uniform, high-density resolution with up to a 3-bit significand, but possesses a limited dynamic range of $\log_2(1.75 / 0.25) \approx 2.81$ binades. Conversely, \textbf{E2M1} sacrifices 1 bit of significand precision to introduce a 2-bit micro-exponent, expanding the local dynamic range to $\log_2(6.0 / 0.5) \approx 3.58$ binades. This broader dynamic-range coverage can be beneficial for activation groups exhibiting larger local magnitude variation.

As illustrated in Figure~\ref{fig:scale_hif4}, for an input block $V_{64} = \{v_1, v_2, \ldots, v_{64}\}$, HIF4 constructs scale factors hierarchically from local peak magnitudes. The 64 elements are first partitioned into 16 four-element groups $\mathcal{G}_i$ ($i \in \{1, \ldots, 16\}$). The local peak magnitude of each group is:
\begin{equation}
V_4[i] = \max_{j \in \mathcal{G}_i} |v_j|.
\end{equation}
Adjacent group peaks are aggregated into eight mid-level peaks:
\begin{equation}
V_8[k] = \max \left( V_4[2k-1], V_4[2k] \right), \quad k \in \{1, \ldots, 8\},
\end{equation}
which finally yield the global block peak $V_{\max} = \max_{k} V_8[k]$.

The hierarchical scale engine maps $V_{\max}$, $V_8$, and $V_4$ into low-bit scale factors. First, the high-precision global scale factor $\mathrm{SF} = \mathrm{clamp}(V_{\max} / \alpha, \mathrm{SF}_{\min}, \mathrm{SF}_{\max})$ is mapped to an E6M2 floating-point representation, where the normalization constant $\alpha$ is set to $7.0$ for S1P2 and $24.0$ for E2M1. The mid-level and fine-level 1-bit scale extensions ($\mathrm{E1}_8$ and $\mathrm{E1}_4$) are evaluated conditionally:
\begin{equation}
\mathrm{E1}_8[k] = \mathbb{I}\left( \frac{V_8[k]}{\mathrm{E6M2}} \ge \beta_1 \right), \quad k \in \{1, \ldots, 8\},
\end{equation}
\begin{equation}
\mathrm{E1}_4[i] = \mathbb{I}\left( \frac{V_4[i]}{\mathrm{E6M2} \cdot 2^{\mathrm{E1}_8[\lceil i/2 \rceil]}} \ge \beta_2 \right), \quad i \in \{1, \ldots, 16\},
\end{equation}
where $\mathbb{I}(\cdot)$ denotes the indicator function, and the dynamic thresholds $(\beta_1, \beta_2)$ are calibrated as $(4.0, 2.0)$ for S1P2 and $(16.0, 8.0)$ for E2M1. Combining these tiers yields the composite element-wise scale $S_{\mathrm{HIF4}}(j) = \mathrm{E6M2} \cdot 2^{\mathrm{E1}_8[\lceil i/2 \rceil] + \mathrm{E1}_4[i]}$ for element $v_j \in \mathcal{G}_i$. The normalized element magnitude $z_j$ is given by:
\begin{equation}
z_j = \frac{|v_j|}{S_{\mathrm{HIF4}}(j)}.
\label{eq:hif4_normalized}
\end{equation}

Finally, $z_j$ is mapped to non-negative discrete formats $\mathcal{F} \in \{\mathcal{F}_S, \mathcal{F}_E\}$:
\begin{equation}
\mathcal{F}_S = \{0,\, 0.25,\, 0.5,\, 0.75,\, 1.0,\, 1.25,\, 1.5,\, 1.75\},
\label{eq:s1p2_format}
\end{equation}
\begin{equation}
\mathcal{F}_E = \{0,\, 0.5,\, 1.0,\, 1.5,\, 2.0,\, 3.0,\, 4.0,\, 6.0\},
\label{eq:e2m1_format}
\end{equation}
with the original sign restored via $\hat{v}_j = \mathrm{sign}(v_j) \cdot S_{\mathrm{HIF4}}(j) \cdot Q_{\mathcal{F}}(z_j)$.

The complete quantization workflow is formalized in Algorithm~\ref{alg:hif4_quant}.

\subsection{Tensor Format Characteristics and Quantization Error Analysis}
\label{sec:format_analysis}

To establish a principled foundation for micro-scaling format assignment, a comprehensive empirical analysis is conducted on the Qwen3-8B architecture. The primary objective is to dissect the intrinsic distributional divergence between weights and activations, and evaluate how local format characteristics impact quantization fidelity both in isolation and within full matrix multiplications.

\subsubsection{Local Dynamic Range Characterization}
\label{sec:dr_profiling}

A diverse dataset comprising 256 sequence tokens is fed into Qwen3-8B. Across all 36 Transformer layers, tensor streams are systematically intercepted at every major GEMM projection boundary:
\begin{enumerate}[leftmargin=*]
    \item \textbf{Self-Attention Projections}: Inputs to the Query ($W_q$), Key ($W_k$), Value ($W_v$), and Attention Output ($W_o$) matrices.
    \item \textbf{FFN / MLP Projections}: Inputs to the Gate ($W_{\text{gate}}$), Up ($W_{\text{up}}$), and Down-projection ($W_{\text{down}}$) matrices.
\end{enumerate}
From these captured streams, $100,000$ 64-element blocks ($V_{64}$) are extracted. Each block is partitioned into 16 fine-grained 4-element sub-groups $\mathcal{G}_i = \{v_{4i-3}, \ldots, v_{4i}\}$.

To quantify local value dispersion within fine-grained groups, we compute the quantization-aware local 4-element Dynamic Range ($\mathrm{DR}_{\mathrm{local}}$) as
\begin{equation}
\mathrm{DR}_{\mathrm{local}}(i) =
\frac{\max_{j \in \mathcal{G}_i} z_j}
{\min_{j \in \mathcal{G}_i,\; z_j \ge 0.1} z_j + \epsilon},
\label{eq:local_dr}
\end{equation}
where $z_j = |v_j| / S_{\mathrm{HIF4}}(j)$ denotes the normalized element magnitude, $S_{\mathrm{HIF4}}(j)$ is the element-wise hierarchical scale defined in Section~\ref{sec:hif4_quantization}, and $\epsilon = 1\times10^{-8}$ prevents division by zero. The denominator is set to $1$ when no element in $\mathcal{G}_i$ satisfies $z_j \ge 0.1$. Specifically, $\mathrm{DR}_{\mathrm{local}}$ measures the normalized peak-to-trough magnitude ratio within four consecutive elements, while filtering out very small normalized magnitudes below $0.1$. A high $\mathrm{DR}_{\mathrm{local}}$ indicates larger intra-group magnitude contrasts, favoring a 4-bit element format with broader local dynamic-range coverage.

\begin{figure}[t]
\centering
\includegraphics[width=0.8\linewidth]{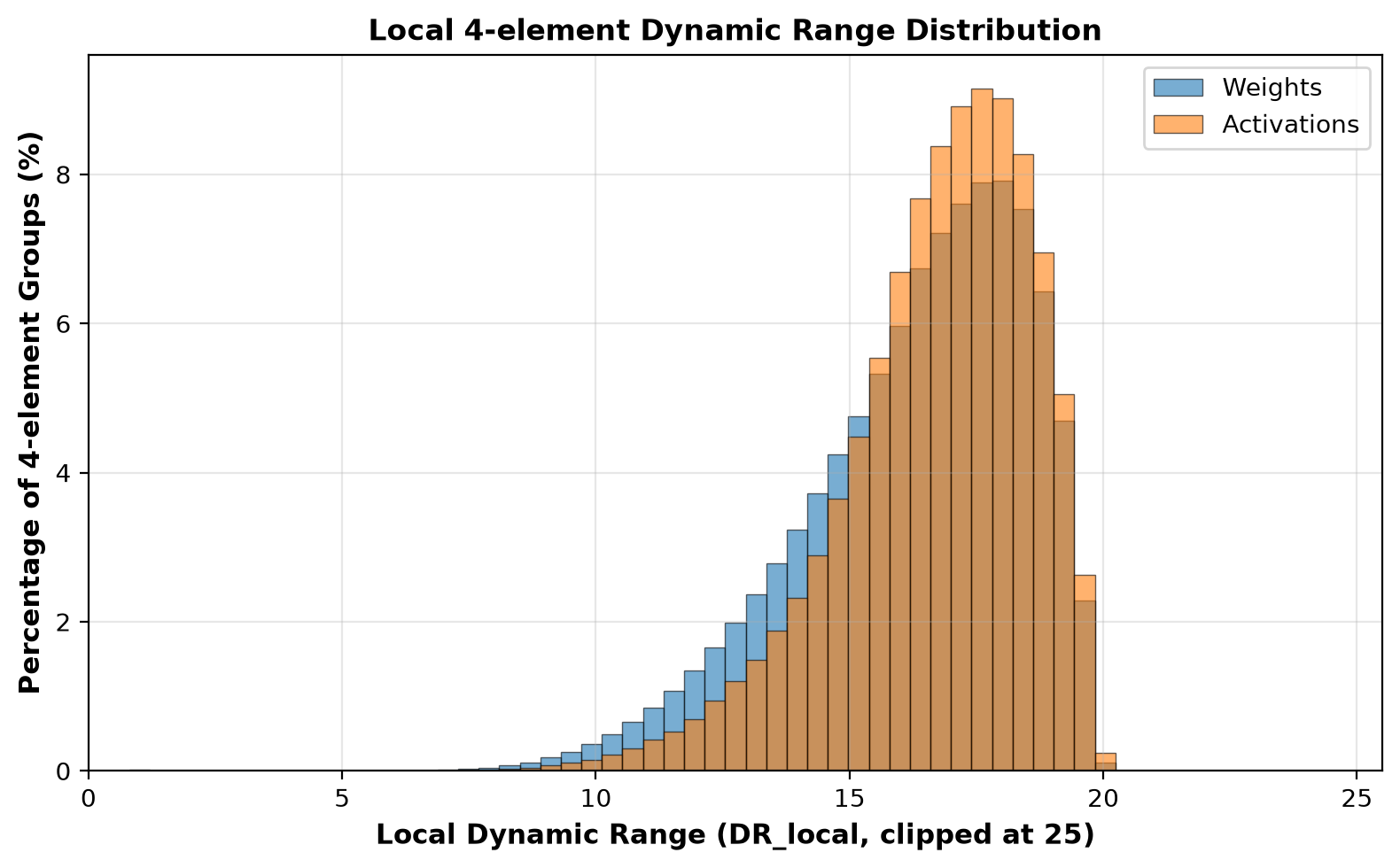}
\caption{Local 4-element Dynamic Range Distribution profiled from Qwen3-8B. The horizontal axis represents the local dynamic range ($\mathrm{DR}_{\mathrm{local}}$, clipped at $25.0$), while the vertical axis denotes the percentage of 4-element groups (\%). Activations display a modest but consistent shift toward higher dynamic range ($\mathrm{Avg\ DR} = 16.70$) compared to weights ($\mathrm{Avg\ DR} = 16.28$).}
\label{fig:local_dr_dist}
\end{figure}

Figure~\ref{fig:local_dr_dist} illustrates the empirical distribution of $\mathrm{DR}_{\mathrm{local}}$ for both weight and activation tensors:
\begin{itemize}[leftmargin=*]
    \item \textbf{Activation Characteristics}: Activation groups shift toward higher dynamic range regimes ($\mathrm{Avg\ DR} = 16.70$), with a larger fraction of groups clustering in higher dynamic range intervals. This reflects the broader local magnitude variation during inference.
    \item \textbf{Weight Characteristics}: Weight groups present a lower average dynamic range ($\mathrm{Avg\ DR} = 16.28$). Although weights span across wide dynamic range boundaries, they exhibit a higher density of low-to-moderate dynamic range groups compared to activations.
\end{itemize}

\subsubsection{Single-Tensor Quantization Error Analysis}
\label{sec:single_tensor_mse}

The divergence in local dynamic range directly governs how weights and activations interact with the S1P2 and E2M1 formats. S1P2 provides high significand precision (up to 3 bits) across a narrow dynamic range ($2.81$ binades), whereas E2M1 dedicates a 2-bit exponent to cover a broader range ($3.58$ binades) at the expense of mantissa resolution.

\begin{figure}[t]
\centering
\begin{subfigure}[b]{0.48\linewidth}
    \centering
    \includegraphics[width=\linewidth]{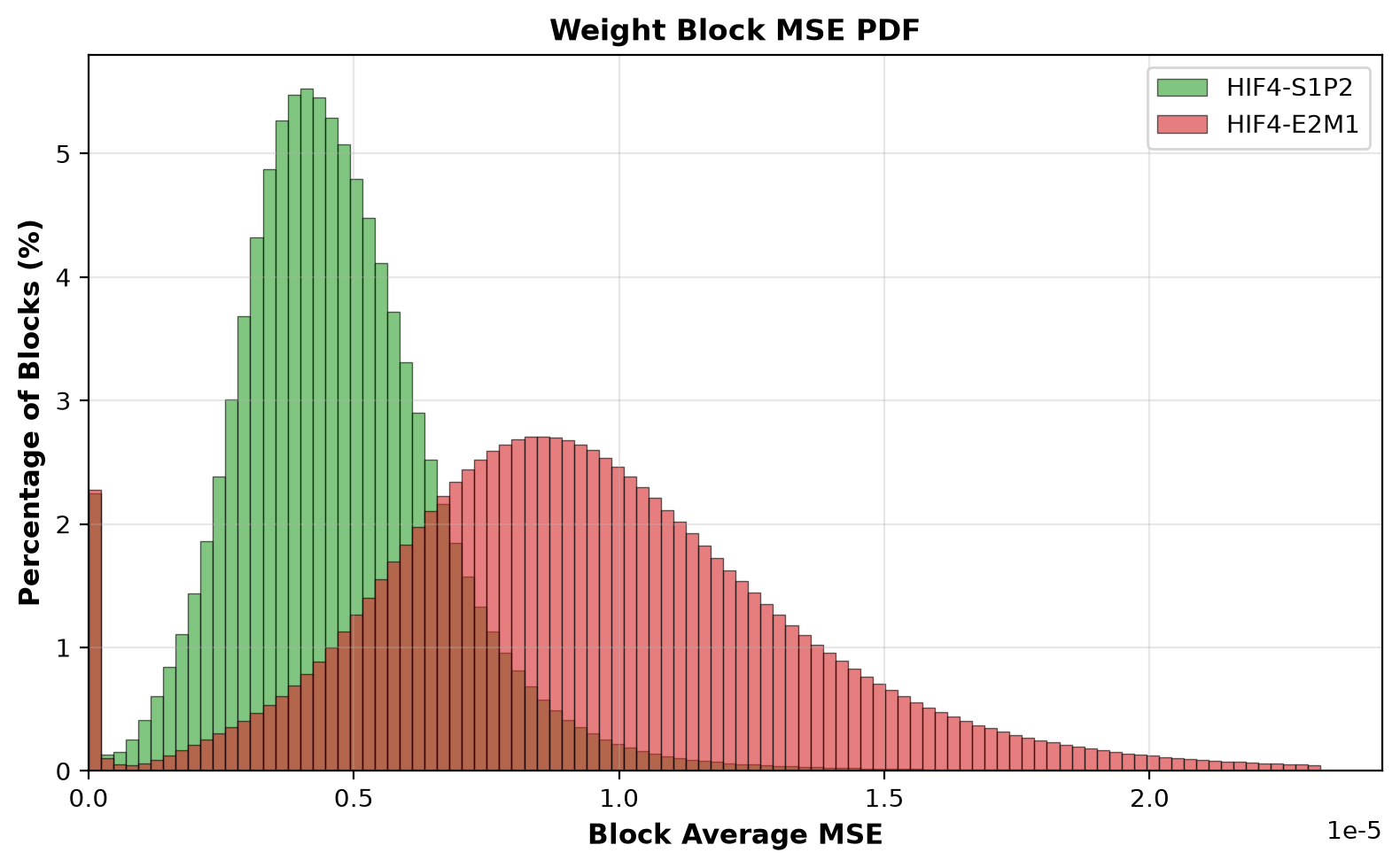}
    \caption{Weight Block MSE Distribution}
    \label{fig:weight_mse_pdf}
\end{subfigure}
\hfill
\begin{subfigure}[b]{0.48\linewidth}
    \centering
    \includegraphics[width=\linewidth]{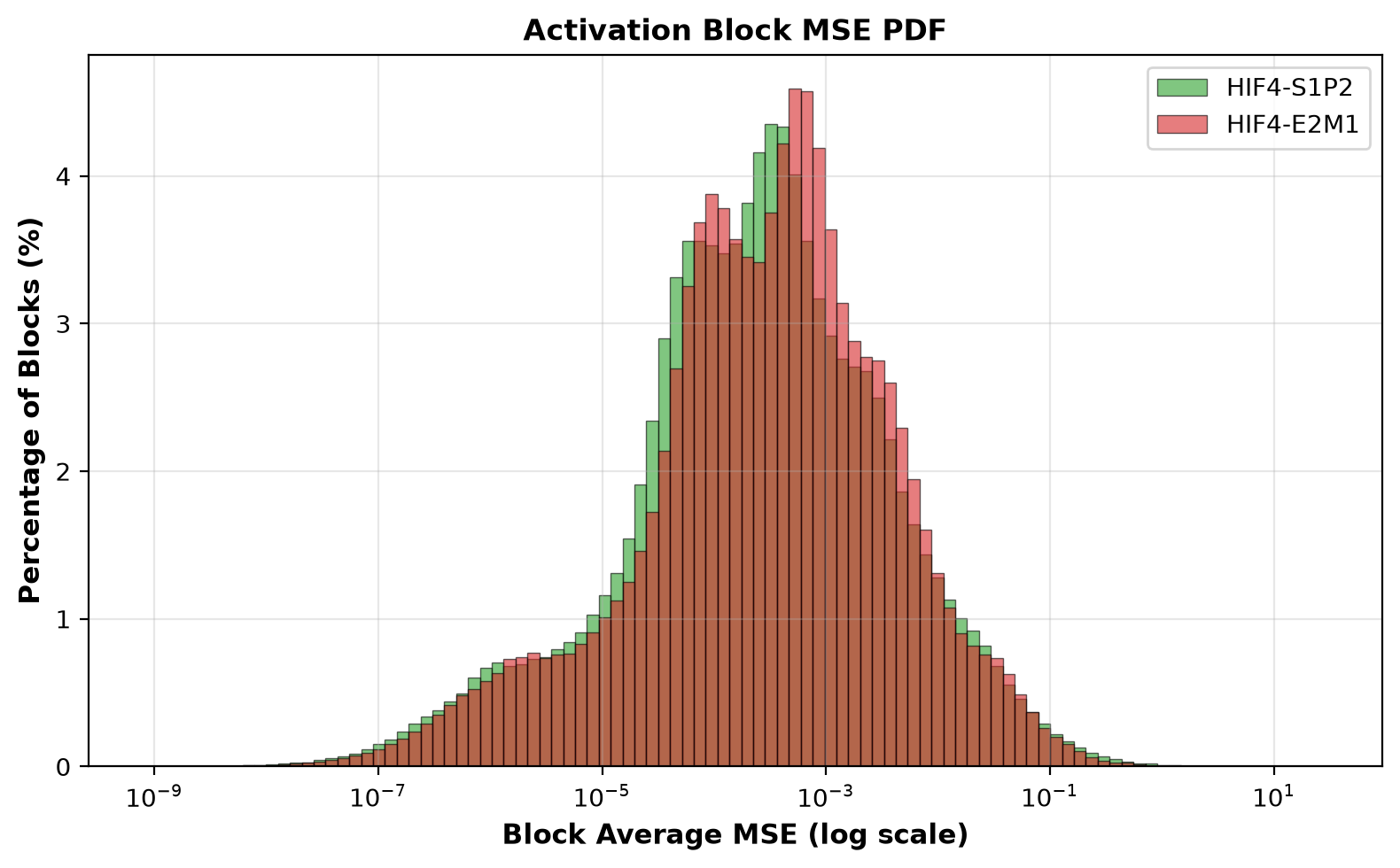}
    \caption{Activation Block MSE Distribution (log scale)}
    \label{fig:act_mse_pdf}
\end{subfigure}
\caption{Quantization MSE empirical distributions for 64-element blocks in Qwen3-8B. 
(a) Weight-block MSE under HIF4-S1P2 and HIF4-E2M1.
(b) Activation-block MSE under the two formats, shown on a logarithmic horizontal axis to expose the broad error range.}
\label{fig:mse_pdf}
\end{figure}

To evaluate quantization fidelity, block-level quantization MSE empirical distributions are plotted in Figure~\ref{fig:mse_pdf}. Figure~\ref{fig:weight_mse_pdf} and Figure~\ref{fig:act_mse_pdf} depict the MSE distributions for weight and activation blocks, respectively. The two distributions use different horizontal-axis scales to reflect their substantially different error ranges. The weight MSE distribution is presented on a linear scale because weight statistics are relatively stable across blocks, resulting in a comparatively concentrated MSE range. In contrast, activation values exhibit larger variation in local dynamic range, which leads to a much wider distribution of block-level MSE values. Therefore, the activation MSE distribution is plotted on a logarithmic horizontal axis to expose the full distribution over multiple orders of magnitude and make both low- and high-error regions observable. The logarithmic scale is used solely for visualization and does not alter the underlying MSE statistics.

For both weights and activations, quantization error is measured against the corresponding unquantized FP16 reference tensor. Given an original tensor $V$ and its quantized-dequantized reconstruction $\hat{V}$, the mean squared error (MSE) is defined as
\begin{equation}
\mathrm{MSE}(V,\hat{V})
=
\frac{1}{N}
\sum_{j=1}^{N}
\left(v_j-\hat{v}_j\right)^2,
\label{eq:mse}
\end{equation}
where $N$ denotes the number of elements in the evaluated tensor or block. As quantified in Table~\ref{tab:single_tensor_mse}, the empirical statistical averages align closely with the local dynamic range characteristics:

\begin{table}[h]
\centering
\caption{Average block-level quantization MSE across Qwen3-8B projections, measured against the unquantized BF16 reference tensors.}
\label{tab:single_tensor_mse}
\begin{tabular}{lccc}
\toprule
\textbf{Tensor Type} &
\textbf{HIF4-S1P2 MSE} ($\downarrow$) &
\textbf{HIF4-E2M1 MSE} ($\downarrow$) &
\textbf{MSE Ratio (S1P2:E2M1)} \\
\midrule
Weight ($W$)
& $\mathbf{4.7179 \times 10^{-6}}$
& $9.5543 \times 10^{-6}$
& $1.00:2.03$ \\
Activation ($A$)
& $5.1681 \times 10^{-3}$
& $\mathbf{4.6095 \times 10^{-3}}$
& $1.12:1.00$ \\
\bottomrule
\end{tabular}
\end{table}

\begin{itemize}[leftmargin=*]
    \item \textbf{Weight Quantization}: HIF4-S1P2 achieves an average MSE of $4.7179 \times 10^{-6}$, whereas HIF4-E2M1 reaches $9.5543 \times 10^{-6}$. Thus, the MSE of HIF4-E2M1 is approximately $2.03\times$ that of HIF4-S1P2, indicating that S1P2 provides substantially lower reconstruction error for the weight distribution. In Figure~\ref{fig:weight_mse_pdf}, the error distribution for S1P2 is visibly shifted toward smaller MSE values. With a higher proportion of low-to-moderate local dynamic-range groups, weights benefit from S1P2's denser representation near zero.

    \item \textbf{Activation Quantization}: Conversely, HIF4-E2M1 achieves an average MSE of $4.6095 \times 10^{-3}$, compared with $5.1681 \times 10^{-3}$ for HIF4-S1P2. Thus, the MSE of HIF4-S1P2 is approximately $1.12\times$ that of HIF4-E2M1, indicating a smaller but consistent advantage of E2M1 for the activation distribution. As shown in the log-scale distribution in Figure~\ref{fig:act_mse_pdf}, E2M1 concentrates more blocks in the mid-MSE range and reduces the extreme high-MSE tail. Although the logarithmic histogram emphasizes the low-to-mid MSE region visually, the average MSE remains sensitive to the high-MSE tail, making the reduced tail particularly relevant to the overall reconstruction error.
\end{itemize}

\subsubsection{Mixed-Format GEMM Quantization Error}
\label{sec:mixed_gemm_mse}

Having verified single-tensor quantization preferences, error accumulation during matrix multiplication $\hat{Y} = Q_A(A) Q_W(W)^T$ is evaluated. End-to-end Output MSE is measured across all four permutations of weight and activation mantissa formats inside the HIF4 scale hierarchy.

\begin{figure}[t]
\centering
\includegraphics[width=0.8\linewidth]{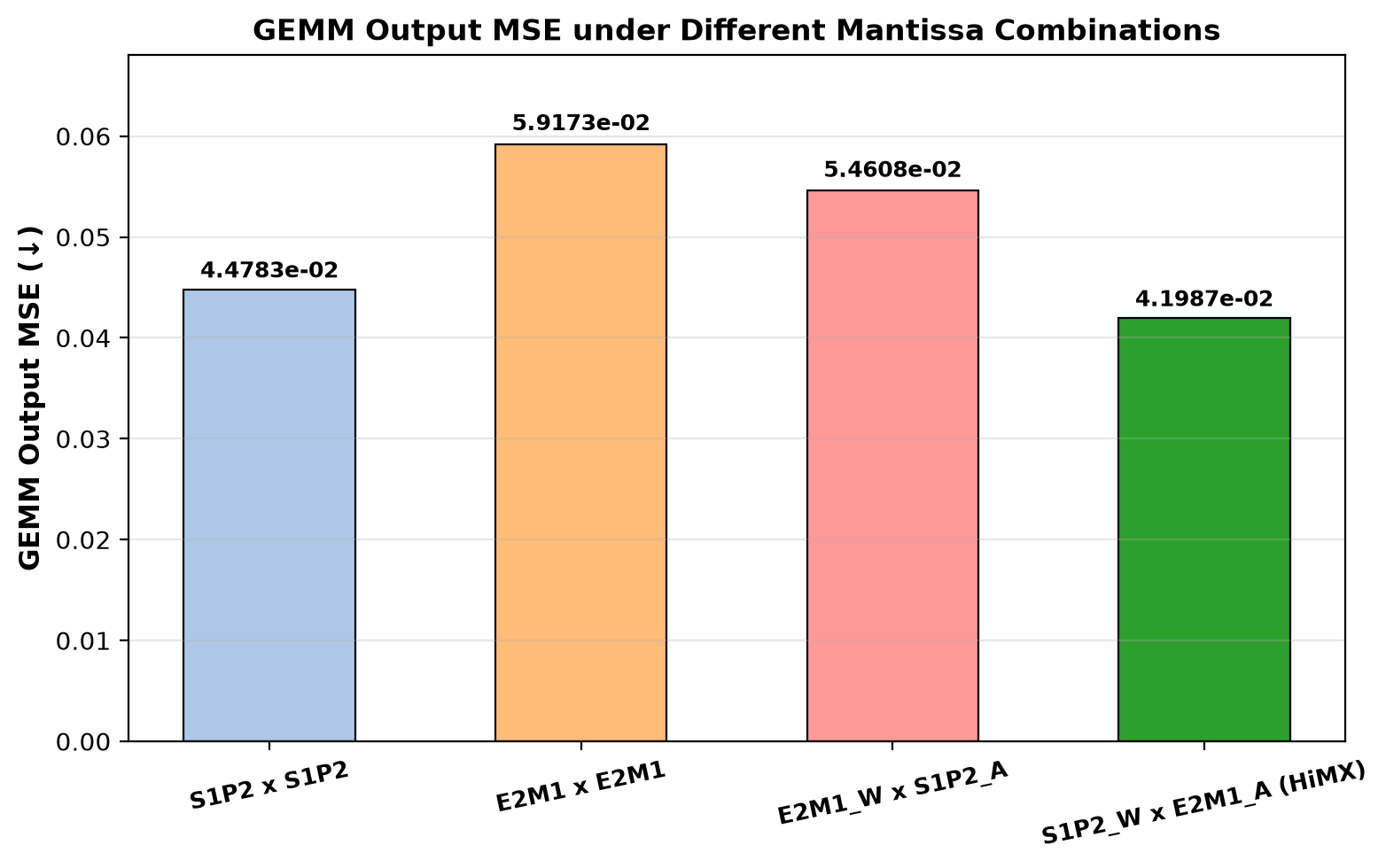}
\caption{GEMM Output MSE under different HIF4 mantissa pairings ($W \times A$). The heterogeneous $\text{S1P2}_W \times \text{E2M1}_A$ pair (HiMX) achieves the lowest matrix multiplication error.}
\label{fig:gemm_output_mse_bar}
\end{figure}


As illustrated in Figure~\ref{fig:gemm_output_mse_bar}, the empirical GEMM reconstruction errors reveal critical design trade-offs:
\begin{enumerate}[leftmargin=*]
    \item \textbf{Homogeneous Baselines}: Pure S1P2 ($\text{S1P2}_W \times \text{S1P2}_A$) yields an Output MSE of $4.4783 \times 10^{-2}$, limited by activation dynamic range truncation. Pure E2M1 ($\text{E2M1}_W \times \text{E2M1}_A$) incurs the highest error ($5.9173 \times 10^{-2}$) due to coarse weight significand resolution.
    \item \textbf{Suboptimal Heterogeneous Pairing}: The inverted pair ($\text{E2M1}_W \times \text{S1P2}_A$) achieves $5.4608 \times 10^{-2}$, failing to resolve the dominant weight quantization error and activation quantization error.
    \item \textbf{Heterogeneous HiMX Design}: The proposed combination ($\text{S1P2}_W \times \text{E2M1}_A$) achieves the optimal Output MSE of $\mathbf{4.1987 \times 10^{-2}}$, delivering a \textbf{6.2\%} error reduction over pure S1P2 and \textbf{29.0\%} over pure E2M1.
\end{enumerate}

These results demonstrate that pairing high-precision weight mantissas ($\text{S1P2}_W$) with wide-dynamic-range activation mantissas ($\text{E2M1}_A$) reduces dot-product error propagation, providing optimal numerical fidelity under a shared hierarchical scaling framework.

\section{HiMX Framework Architecture}
\label{sec:himx_arch}

\begin{figure}[t]
\centering
\begin{subfigure}[b]{0.48\linewidth}
\centering
\includegraphics[width=\linewidth]{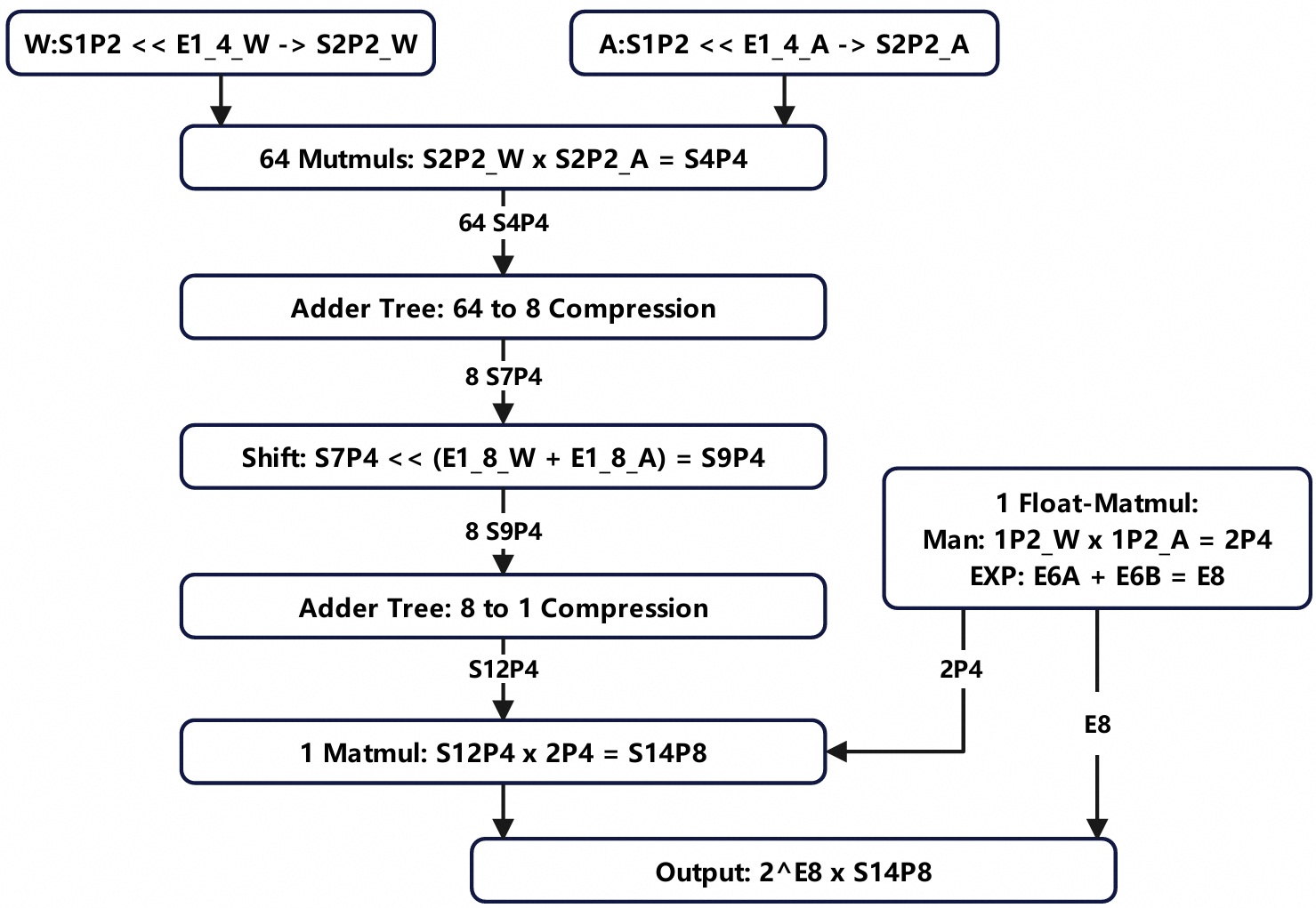}
\caption{HIF4-S1P2 architecture\cite{hif4}.}
\label{fig:datapath_s1p2}
\end{subfigure}
\hfill
\begin{subfigure}[b]{0.48\linewidth}
\centering
\includegraphics[width=\linewidth]{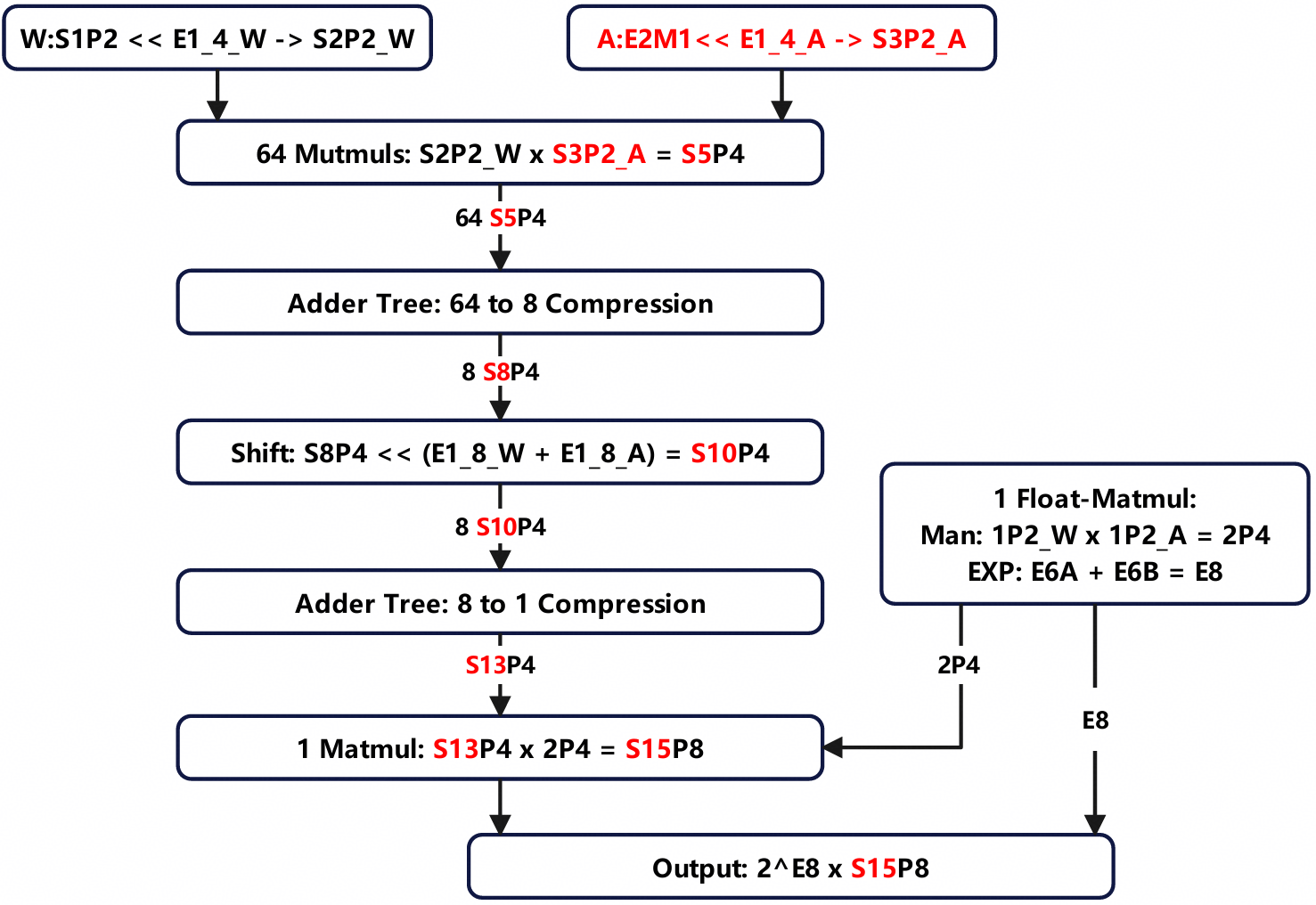}
\caption{HiMX S1P2$\times$E2M1 architecture.}
\label{fig:datapath_himx}
\end{subfigure}
\caption{
Comparison of the 64-element dot-product architecture between
(a) the original HIF4-S1P2 configuration and
(b) the proposed HiMX S1P2$\times$E2M1 configuration.
Both datapaths preserve the three-level HIF4 hierarchical scaling and hierarchical accumulation structure, while HiMX introduces a wider integer representation for the E2M1 activation operand and the corresponding multiplication datapath.
}
\label{fig:datapath}
\end{figure}

HiMX extends the original HIF4 computation architecture to support heterogeneous mantissa formats for weights and activations while preserving the hierarchical scaling and integer-based dot-product computation of HIF4. Figure~\ref{fig:datapath} compares the original HIF4-S1P2 datapath with the proposed heterogeneous S1P2$\times$E2M1 datapath. The key architectural modification is localized to the operand representation and the corresponding multiplication stage: the hierarchical scale processing, partial-sum compression, and final global-scale reconstruction remain structurally consistent with the original HIF4 design.

Figure~\ref{fig:datapath_s1p2} illustrates the original HIF4-S1P2 computation flow. Within each 64-element HIF4 block, the third-level scale extensions are first incorporated into the local S1P2 mantissas. The resulting integer-compatible operands are then processed by 64 parallel multiplications. The products are compressed through a hierarchical adder tree, where the first reduction stage aggregates 64 products into 8 partial sums. The second-level scale information is subsequently introduced through shift operations, avoiding the need for general-purpose multipliers for the hierarchical scaling factors. A second adder-tree stage then reduces the 8 shifted partial sums into a single accumulated result. Finally, the accumulated integer value is combined with the remaining global scale information to reconstruct the output. This hierarchical organization allows most of the 64-element dot product to remain in the integer domain and is one of the main reasons for the computational efficiency of HIF4\cite{hif4,msfp,shared_microexp}.

HiMX retains this overall computation flow but changes the representation of the activation operand, as shown in Fig.~\ref{fig:datapath_himx}. The weight path remains unchanged and uses the original HIF4-S1P2 representation. In contrast, the activation path adopts the HIF4-E2M1 format. After incorporating the fine-level $\mathrm{E1}_4$ scale information (the sixteen 1-bit fine-level scales introduced in Section~\ref{sec:hif4_quantization}), the weight and activation operands are represented as $\mathrm{S2P2}_W$ and $\mathrm{S3P2}_A$, respectively. Here, $\mathrm{S}x\mathrm{P}y$ denotes a sign bit followed by an $x$-bit integer part and a $y$-bit fractional part in the internal fixed-point representation. Compared with the original S1P2$\times$S1P2 datapath, the activation operand therefore requires one additional bit in its integer representation. The multiplication stage is consequently expanded from $\mathrm{S2P2}\times\mathrm{S2P2}$ to $\mathrm{S2P2}\times\mathrm{S3P2}$, resulting in a wider intermediate product.

This modification is important because the E2M1 format provides a substantially wider local numerical range than S1P2. Instead of forcing both operands to use the same mantissa representation, HiMX preserves the denser S1P2 representation for weights while providing the activation operand with the additional representation capacity required by E2M1. The resulting datapath therefore matches the numerical characteristics of the two operands more closely. In particular, the wider E2M1 representation allows activation values with larger local dynamic ranges to be represented without sacrificing the hierarchical scaling structure of HIF4. At the same time, the weight path does not incur the additional representation cost of E2M1, since the empirical analysis in Section~\ref{sec:analysis} shows that S1P2 provides substantially lower quantization error for the weight distribution.

After the heterogeneous multiplication stage, HiMX preserves the original HIF4 hierarchical accumulation strategy. The 64 multiplication results are first compressed into 8 partial sums. The combined $\mathrm{E1}_8$ scale information of the weight and activation operands is then implemented through integer shifts, after which a second adder tree produces the final integer accumulation result. The global E6M2 scale factors are reconstructed at the output stage in the same manner as HIF4. Consequently, the introduction of heterogeneous mantissa formats does not require a new hierarchical accumulation architecture. Instead, the additional complexity is concentrated primarily in the operand conversion and multiplication stage.

The main hardware overhead of HiMX therefore comes from the increased integer bit-width of the activation operand in the heterogeneous multiplication. Compared with the original S1P2$\times$S1P2 datapath, the E2M1 activation operand requires a wider integer representation (S3P2 instead of S2P2), which increases the area of the corresponding multiplication units. A straightforward implementation that introduces dedicated arithmetic resources for the wider heterogeneous operation would incur a noticeable resource overhead. However, directly duplicating a complete multiplication datapath is unnecessary because the original HIF4 computation already contains wide arithmetic units that can be reused.

To reduce this overhead, one possible implementation is to reuse the existing wider arithmetic resources of the HIF4 datapath through sub-word parallelism\cite{mxplus,m2xfp}, rather than assigning independent multiplication hardware to the additional E2M1 path. Using FP16 as an example, the 11-bit mantissa path of an existing FP16 multiplier can be partitioned into two disjoint sub-word fields, each performing an independent 4-bit multiplication. Since each wide mantissa multiplier can host two independent 4-bit sub-word multiplications, the existing arithmetic arrays can be reused to reconstruct a larger number of FP4 multiplication operations without adding new multiplier units. More generally, modern accelerators already provide native support for multiple narrow numerical formats such as FP8 and MXFP4\cite{nvfp4,ocp_mx,mxplus}, and the corresponding arithmetic resources can similarly be reused for the heterogeneous E2M1 path. The two 4-bit sub-words occupy disjoint bit fields of the wider operand, and the partial products generated from different sub-words are isolated by the corresponding bit-field alignment, avoiding cross-term contamination. The outputs are then extracted from the corresponding bit ranges after accumulation. Under such a resource-sharing scheme, the estimated additional area of the multiplication array is approximately $2\%$--$5\%$ of the combined FP16 and incremental FP4 resources, with the overhead dominated by the additional alignment and masking logic rather than by new multiplier units.

Overall, the architecture provides a favorable trade-off between numerical flexibility and hardware efficiency. HiMX allows S1P2 and E2M1 to be assigned independently according to the statistical characteristics of weights and activations, while retaining the original HIF4 hierarchical scaling and accumulation mechanism. The heterogeneous representation improves the ability to match operand-specific dynamic-range requirements, whereas SIMD-based resource sharing limits the resulting hardware overhead to approximately $2\%$--$5\%$. This design demonstrates that operand-level mantissa heterogeneity can be introduced into HIF4 without fundamentally changing its integer-oriented dot-product architecture.

\section{Experiments}
\label{sec:exp}

We evaluate HiMX on the Qwen3 family of large language models\cite{qwen3_report}, including Qwen3-8B and Qwen3-30B-A3B, covering six standard benchmarks: MMLU, GSM8K, HumanEval, MBPP, ARC-Challenge (ARC-C), and ARC-Easy (ARC-E)\cite{mmlu,gsm8k,humaneval,mbpp,arc}. We compare four HIF4 mantissa configurations: homogeneous baselines ($\text{S1P2}_W \times \text{S1P2}_A$, $\text{E2M1}_W \times \text{E2M1}_A$) and heterogeneous pairs ($\text{E2M1}_W \times \text{S1P2}_A$, $\text{S1P2}_W \times \text{E2M1}_A$), with FP16 $\times$ FP16 serving as the full-precision reference.

\begin{table}[t]
\centering
\small
\setlength{\tabcolsep}{3.5pt}
\caption{
End-to-end accuracy of Qwen3 models under different HIF4 mantissa configurations.
The Mean column reports the average accuracy over six benchmarks:
MMLU, GSM8K, HumanEval, MBPP, ARC-C, and ARC-E.
Values in parentheses denote the accuracy drop relative to the FP16 baseline.
}
\label{tab:qwen3_himx}
\begin{tabular}{c c c c c c c c c}
\toprule
\textbf{Model} &
\textbf{Weight--Activation} &
\textbf{MMLU} &
\textbf{GSM8K} &
\textbf{HumanEval} &
\textbf{MBPP} &
\textbf{ARC-C} &
\textbf{ARC-E} &
\textbf{Mean} \\
\midrule

\multirow{5}{*}{Qwen3-8B}
& FP16 $\times$ FP16
& 74.968 & 88.476 & 64.024 & 65.000 & 67.577 & 87.963
& 74.668 \\

& S1P2 $\times$ S1P2
& 73.159 & 85.140 & 59.146 & 63.200 & 65.529 & 87.542
& 72.286 (-2.382) \\

& E2M1 $\times$ E2M1
& 72.483 & 86.429 & 59.756 & 61.800 & 63.481 & 87.416
& 71.894 (-2.774) \\

& E2M1 $\times$ S1P2
& 72.710 & 85.747 & 60.366 & 62.600 & 64.334 & 86.911
& 72.111 (-2.557) \\

& S1P2 $\times$ E2M1
& 72.547 & 86.277 & 63.415 & 61.800 & 65.785 & 87.584
& \textbf{72.901 (-1.767)} \\

\midrule

\multirow{5}{*}{Qwen3-30B-A3B}
& FP16 $\times$ FP16
& 82.033 & 89.765 & 76.220 & 76.400 & 72.867 & 90.572
& 81.310 \\

& S1P2 $\times$ S1P2
& 80.822 & 88.855 & 70.732 & 73.200 & 71.075 & 89.773
& 79.076 (-2.234) \\

& E2M1 $\times$ E2M1
& 79.868 & 90.523 & 71.951 & 73.200 & 69.454 & 89.773
& 79.128 (-2.182) \\

& E2M1 $\times$ S1P2
& 80.352 & 88.462 & 72.171 & 74.800 & 70.478 & 89.899
& 79.360 (-1.950) \\

& S1P2 $\times$ E2M1
& 80.730 & 88.704 & 71.950 & 74.400 & 70.734 & 90.236
& \textbf{79.459 (-1.851)} \\

\bottomrule
\end{tabular}
\end{table}

As summarized in Table~\ref{tab:qwen3_himx}, the heterogeneous HiMX configuration ($\text{S1P2}_W \times \text{E2M1}_A$) achieves the highest mean accuracy on both Qwen3-8B and Qwen3-30B-A3B. It reaches $72.901$ on Qwen3-8B and $79.459$ on Qwen3-30B-A3B, corresponding to accuracy losses of $1.767$ and $1.851$ percentage points from the respective FP16 baselines. Compared with the homogeneous ($\text{S1P2}_W \times \text{S1P2}_A$) configuration, the heterogeneous configuration improves the mean accuracy by $0.615$ and $0.383$ percentage points on Qwen3-8B and Qwen3-30B-A3B, respectively. It also outperforms the homogeneous ($\text{E2M1}_W \times \text{E2M1}_A$) configuration and the inverted heterogeneous pairing ($\text{E2M1}_W \times \text{S1P2}_A$) on both models. We note that the mean advantage of HiMX is not uniform across every individual benchmark: on some benchmarks the homogeneous or inverted heterogeneous configurations remain slightly better, and the improvement of HiMX mainly comes from its consistently better average behavior.

These results are consistent with our tensor-level analysis, suggesting that heterogeneous mantissa assignment can provide a more favorable balance between local representation precision and dynamic-range coverage than a single mantissa format. In particular, assigning S1P2 to weights and E2M1 to activations provides the best end-to-end accuracy among the evaluated configurations. The consistent improvement over the homogeneous S1P2 baseline demonstrates the practical benefit of introducing operand-level mantissa heterogeneity within the HIF4 framework.

From a hardware perspective, the dominant additional overhead of HiMX stems from the increased integer bit-width of the activation operand in the heterogeneous multiplication. By converting heterogeneous mantissas into fixed-point integer operands with operand-specific widths and by reusing the existing SIMD arithmetic resources through sub-word packing, the estimated additional area of the multiplication array is limited to approximately $2\%$--$5\%$ of the combined FP16 and incremental FP4 resources, rather than scaling proportionally with independent format-specific datapaths.


\section{Conclusion}
\label{sec:conclusion}

We presented HiMX, a mixed-format multiplication framework for HIF4 quantization and efficient low-bit LLM inference. The key observation behind our approach is that a single 4-bit mantissa format is not uniformly optimal across different LLM tensors. HIF4-S1P2 provides dense and relatively uniform local representation, making it well suited to concentrated weight distributions, whereas HIF4-E2M1 provides a wider non-uniform format that can better accommodate tensors with larger local dynamic-range requirements.

Based on this observation, HiMX assigns different mantissa formats to different operands while retaining the hierarchical HIF4 scaling framework. To efficiently support heterogeneous operands, HiMX maps each mantissa format to a fixed-point integer operand and aligns the operand widths at the multiplier input, enabling different mantissa representations to share a unified integer multiplication array. This design confines the additional overhead primarily to the increased integer bit-width of the E2M1 activation operand and to the conversion and alignment stages, and the SIMD-based resource reuse limits the additional multiplication-array area to approximately $2\%$--$5\%$ of the combined FP16 and incremental FP4 resources.

Our numerical and end-to-end evaluations demonstrate the effectiveness of this mixed-mantissa design. Block-level analysis shows that the choice of mantissa format affects both weight and activation quantization fidelity, while GEMM-level evaluation identifies ($\text{S1P2}_W \times \text{E2M1}_A$) as the most effective configuration among the evaluated combinations. On Qwen3-8B, this configuration achieves a mean accuracy of $72.901$ across six benchmarks, improving over the homogeneous ($\text{S1P2}_W \times \text{S1P2}_A$) baseline by $0.615$ percentage points and remaining within $1.767$ percentage points of the FP16 baseline. On Qwen3-30B-A3B, it achieves a mean accuracy of $79.459$, improving over the homogeneous S1P2 configuration by $0.383$ percentage points with an FP16 accuracy loss of $1.851$ percentage points.

Overall, these results show that introducing operand-level mantissa heterogeneity within the HIF4 framework can improve low-bit LLM accuracy without requiring independent multiplication datapaths for different formats. Future work will investigate distribution-aware format selection based on the characteristics of different models and tensor types, explore a broader range of low-bit numerical formats and bit widths, and further reduce the hardware and implementation overhead associated with supporting heterogeneous formats. We will also investigate the integration of distribution-aware format selection with other LLM compression techniques, including sparsity, KV-cache compression, and MoE-aware quantization.

\bibliographystyle{unsrt}  
\bibliography{references}

\end{document}